\documentclass[sigconf]{acmart}

\usepackage{booktabs} 
\usepackage{subcaption}

\setcopyright{rightsretained}

\acmDOI{10.475/123_4}

\acmISBN{123-4567-24-567/08/06}

\acmConference[WOODSTOCK'97]{ACM Woodstock conference}{July 1997}{El
  Paso, Texas USA}
\acmYear{1997}
\copyrightyear{2016}

\acmArticle{4}
\acmPrice{15.00}

\editor{Jennifer B. Sartor}
\editor{Theo D'Hondt}
\editor{Wolfgang De Meuter}

\begin{document}
\title{Efficient GAN-based method for cyber-intrusion detection}

\author{Hongyu Chen}

\affiliation{
  \institution{Shanghai Jiao Tong University, China}
}

\author{Li Jiang}

\affiliation{
  \institution{Shanghai Jiao Tong University, China}
}

\begin{abstract}
Ubiquitous anomalies endanger the security of our system constantly. They may bring irreversible damages to the system and cause leakage of privacy. Thus, it is of vital importance to promptly detect these anomalies. Traditional supervised methods such as Decision Trees and Support Vector Machine~(SVM) are used to classify normality and abnormality. However, in some case the abnormal status are largely rarer than normal status, which leads to decision bias of these methods. Generative adversarial network~(GAN) has been proposed to handle the case. With its strong generative ability, it only needs to learn the distribution of normal status, and identify the abnormal status through the gap between it and the learned distribution. Nevertheless, existing GAN-based models are not suitable to process data with discrete values, leading to immense degradation of detection performance. To cope with the discrete features, in this paper, we propose an efficient GAN-based model with specifically-designed loss function. Experiment results show that our model outperforms state-of-the-art models on discrete dataset and remarkably reduce the overhead.
\end{abstract}

%
%


\settopmatter{printacmref=false}
\renewcommand\footnotetextcopyrightpermission[1]{}
\pagestyle{plain}

\maketitle

\section{Introduction}

The vicious cyber-intrusions endanger our devices all the time, they have many severe consequences such as the unauthorized divulgement of information, the tampering, destruction, and expungement of data. Thus unsupervised and efficacious detection is required to respond to these malicious intrusions against networks and computers. Lots of explorations have been done with both statistical learning methods and neural networks\cite{a,b}.

To effectively classify the normal internet connections and intrusions. Classical supervised methods such as Decision Trees\cite{e,f,57} and SVM\cite{98} are applied to this detection task. However, the number of anomalous samples are usually largely fewer than that of normal samples. The obvious imbalance between the amount of normal and anomalous samples is fatal to these statistical learning models. Because the imbalance may bring a bias on the judgement of these supervised models. Besides, when a new kind of appears, it will be time-costing to manually give the label.

Other unsupervised methods such as Local Outlier Factor\cite{lof}, Robust Covariance\cite{rc}, Isolation Forest\cite{if}, etc., can mitigate the above issues to some degree. They detect the abnormal samples mainly based on the density of probability. In other words, they depend on the occurrence frequency of the samples. These methods have better performance on imbalanced dataset compared with supervised methods. However, misjudgment will ineluctably happen if a non-emerging normal sample deviates from the centralized region. As Figure 1 shows, those frequency-based methods need sufficient capacity to learn the features of normal samples. Unfortunately, in the real world, the data usually has complex feature distribution, which limits by the feature representation ability of these frequency-based methods.

\begin{figure}
\includegraphics[width=8cm,height=6cm]{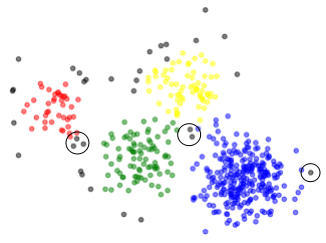}
\caption{LOF on anomaly detection. Chromatic points are different types of normal samples. Black points are judged anomalous samples. The black points in black circle are misjudged normal samples due to their low frequency of occurrence and deviation from centralized region.}
\vspace{-8pt}
\end{figure}

In recent years, generative adversarial network(GAN)\cite{gan} is prevalent for its strong generative ability. In its classical model, a \textit{generator} and a \textit{discriminator} are trained to generate results in an adversarial way -- \textit{generator} tries to generate samples that can fit the distribution of real samples, and the \textit{discriminator} tries to distinguish the generated samples~(fake samples) from the real samples. Since GAN is able to learn the distribution of data, it can naturally be used to learn the distribution of normal data, especially where anomalies are scarce in the training set. In testing~(shown as Figure 2), we can find the most similar sample with the testing sample from the learned distribution and through the defined \textit{anomaly score} based on the intensity of the discrepancy between the testing samples and the found samples we can know how anomalous the testing sample is.

\begin{figure}
\includegraphics[width=6cm,height=4.7cm]{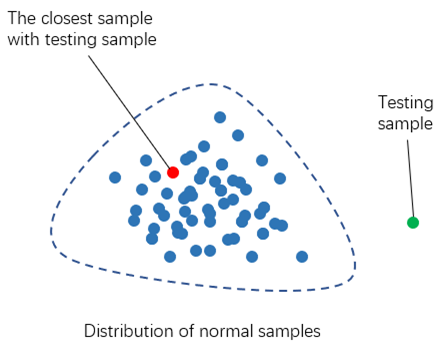}
\caption{Green point is the tesing sample. Red point is the most similar sample with the testing sample a generator can generate, which can also seen as the 'normal version' of testing sample.}
\vspace{-8pt}
\end{figure}

The essence of using GAN for anomaly detection is learning the feature of normal data and accurately find the 'normal version' of the testing sample. AnoGAN\cite{anogan} is proposed to better extract the feature of normal samples, by establishing a mapping between the real space and latent space. And through the discrepancy between the mapped sample and testing sample, we can approach the optimal 'normal version' of testing sample. But this mapping is based on the back-propagation algorithm, thus when the dimension of data increases, this model will be time-consuming and not suitable for timely intrusion detection. To ease the above challenge, a new model~\cite{egan} is adopted, inspired by the structure of BiGAN\cite{bigan}, to remarkably reduce the time cost.

However, the following problems remains in the anomaly detection task. In the training data for cyber-intrusion detection, those discrete features are lethal to traditional GANs, during whose training process the loss criterion is cross-entropy -- a measurement which is not proper to weigh two distribution without overlapping.

To overcome the above hurdles, in this paper, we proposed a GAN-based model with refined loss function to obtain an outstanding performance on the imbalanced dataset with discrete features. Furthermore, we used the \textit{multiple intermediate layers} to soften the decision of \textit{discriminator} to obtain a more moderate result. 

The remainder of this paper is organized as follows: we describe the related work in Sec~\ref{sect:sample-relatedwork}. Sec~\ref{sect:sample-ourmodel} expatiates the details of our model. We show the experiments in Sec~\ref{sect:sample-experiment}.
The paper is concluded in Sec~\ref{sect:sample-conclusion}.

\section{Related Work}\label{sect:sample-relatedwork}
\subsection{Traditional Methods}
Copious work had been done in cyber-intrusion detection, as surveyed by Anna L. Buczak and Erhan Guven\cite{a}. Blowers et al. adopted clustering method like DBSCAN\cite{c} to distinguish the anomalous intrusion, Khan\cite{d} used Genetic Algorithm to detect intrusion. In addition, decision trees such as ID3\cite{e} and C4.5\cite{f} algorithm also be applied to the detection task. Li et al. proposed SVM classifier with RBF kernel to mark off the intrusion. Nevertheless, these methods all have deficiency respectively, for instance, decision trees need enough memory to run, so they may improper for a too big dataset. Furthermore, SVM-involved methods usually need an optimal hyperplane to divide the normal features and abnormal features, which implies they are time-consuming approaches to high-dimension data. Meanwhile, as mentioned before, the scarce anomalous sample in training also leads to their underperformance.

\subsection{GAN-based methods}
The initial purpose to apply GAN on anomaly detection task is to learn the distribution of normal status; then through the discrepancy between the testing sample and learned distribution, we can judge whether the testing sample is in anomalous status(suffered from a cyber-intrusion). However, the problem is how to evaluate the discrepancy after the learning about the distribution of normal status is completed?

Thomas Schlegl et al.\cite{anogan} explained since the mapping from latent space to real space(the task of \textit{generator}) is well learned, the result generated by \textit{generator} should perfectly fit the distribution of normal status. Hence if the corresponding latent status of a testing sample is found, then through the \textit{generator}, the latent status can be mapped into real space and the regenerated sample, which also can be seen as a 'normal version' of the testing sample, should fit the distribution of normal status. However, the basic structure of GAN only unilaterally reflects the latent space into real space, inversely finding the corresponding latent status of the testing sample in real space is challenging. Based on smooth transition of latent space\cite{dcgan} that two status close in latent space generates two similar samples in real space, Thomas Schlegl et al. randomly chose a latent status $z_1$ in latent space at the beginning, then obtained $G(z_1)$, a real space sample, through \textit{generator}. Finally, through the \textit{anomaly score}(a derivable loss function) defined between $G(z_1)$ and testing sample, the location of the corresponding latent status of the testing sample is optimized by an iterative process via back-propagation algorithm. 

However, the back-propagation steps are time-consuming and not suitable for timely cyber-intrusion detection. In later work, Houssam Zenati et al.\cite{egan} adopted BiGAN\cite{bigan} to simultaneously learn the mapping from real space to latent space through \textit{encoder} when the mapping from latent space to real space was learned by \textit{generator}. The advent of \textit{encoder} remarkably reduces the time cost in finding the latent status.

As for cyber-intrusion detection task, the two models above still have deficiency -- the discreteness in the features of training data is fatal to their cross-entropy loss function during the training process(Sec~\ref{sect:anoass}), thus causes \textit{mode collapse}\cite{tog}. When that happens, all input samples will be mapped to similar output samples; and the optimization fails to make progress~\cite{cyclegan}. This is a vital deficiency because when training process is trapped in this situation, we can simply extract some localized features. What we learned from these features is just an incomplete distribution~(several local onefold parts from the distribution of normal status). Consequently, an incomplete distribution can hardly dispose the the diversification of cyber-intrusion. 

To get out of this plight, we refined the loss function during training process of the two previous models to have a better performance on cyber-intrusion detection(Sec~\ref{sect:training}) and ameliorated the \textit{anomaly score} to effectively extract features in deeper networks(Sec~\ref{sect:anoass}).

\section{Our Model}\label{sect:sample-ourmodel}
This section first describes the structure of proposed GAN model; it then illustrates the training process of the model. The last part is about the anomaly assessment we designed.



\subsection{Proposed GAN Structure}
The skeleton of our model is derived from BiGAN\cite{bigan}, which not only reflects the latent samples into real samples in \textit{generator} but also synchronously reflects the real samples into their latent status through \textit{encoder} showed in Figure 3. The addition of the \textit{encoder} is meaningful because with this mechanism there is no need to find the corresponding latent status of a sample anew in later testing part, which process will cost much time based on back-propagation method. In BiGAN, when we learning the mapping from latent space to real space, the inverse mapping from real space to latent space will simultaneously be learned by \textit{encoder}. Benefited from this ingenious structure, we can instantly obtain the corresponding reflection in latent space belongs to a certain testing sample through the learned mapping. Apart from time-saving, the bilateral constraint is also conducive to more effective feature extraction and more unambiguous mapping\cite{bigan}.

\begin{figure}
\includegraphics[width=8cm,height=4.2cm]{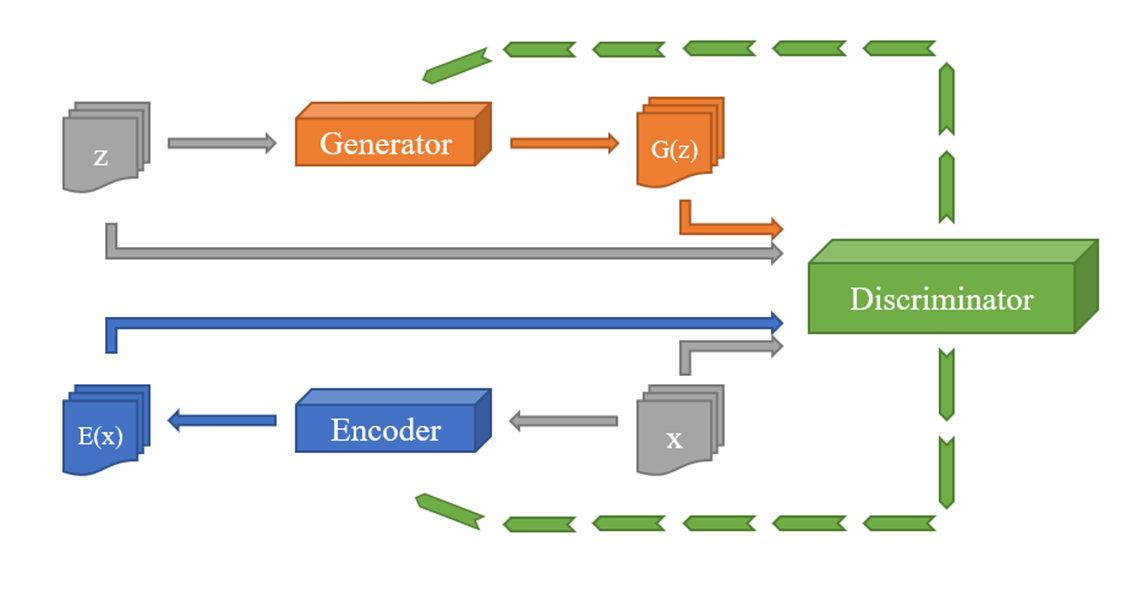}
\caption{Structure of BiGAN. z and E(x) are in latent space, G(z) and x are in real space. After training process commences, z~(initialized latent status) and x~(samples from training set) are converted into G(z) and E(x) respectively, then the two pairs (G(z), z) and (x, E(x)) will be stuffed into \textit{discriminator}, finally the gradient update will be send back to optimize the \textit{generator} and the \textit{encoder}.}
\vspace{-8pt}
\end{figure}

\subsection{Training procedure}\label{sect:training}
Different strategies had been explored to optimize the training process of GAN. The most widely used goal is a minimax objective which illustrated as below:
\begin{align}
    \min\limits_{G}\max\limits_{D}V(D,G)
\end{align}
\begin{align}
    V(D,G)=& \;\mathbb{E}_{x\sim pX} [\log D(x)] + \mathbb{E}_{z\sim pZ} [\log(1-D(G(z)))]
\end{align}
Where D, G separately represents \textit{discriminator} and \textit{generator}, $pX$ is the distribution of normal samples $x$, $pZ$ is the distribution over the latent space.

\begin{figure*}
\includegraphics[width=14cm,height=6cm]{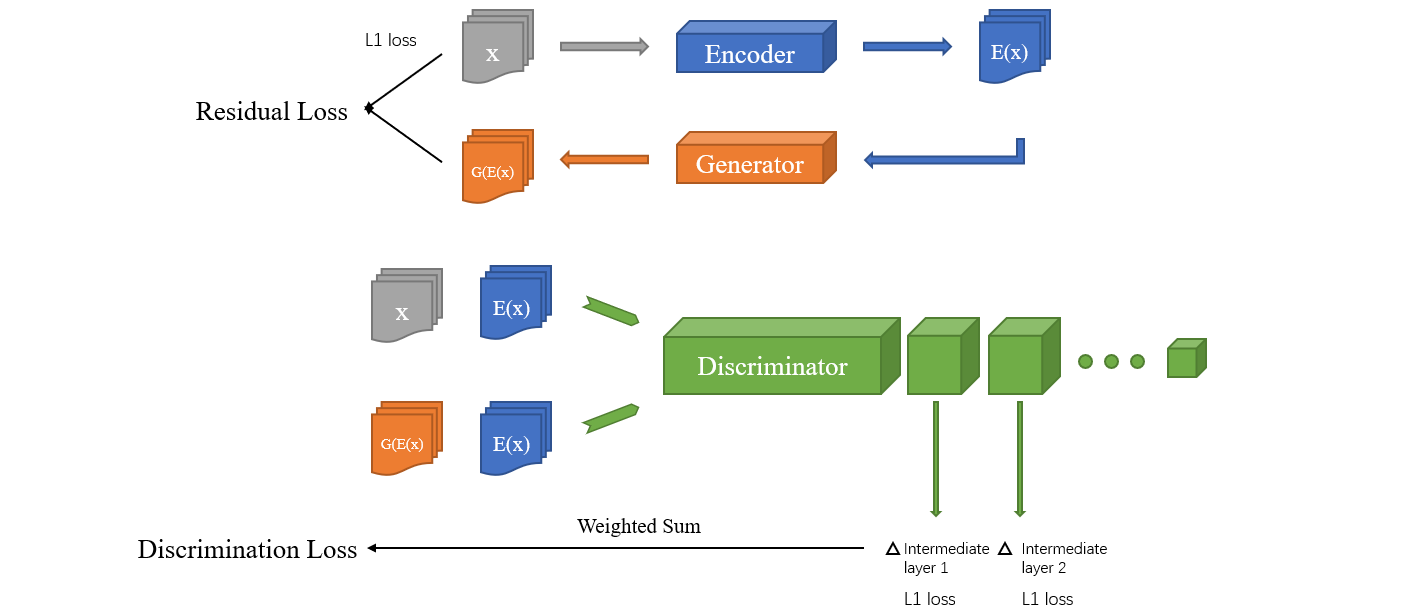}
\caption{In testing, when a testing sample x comes in, its corresponding latent distribution E(x) will be found through \textit{Encoder} and through \textit{Generator} the 'normal version', G(E(x)), of testing sample will be found. Their L1 loss is residual loss. After they are fed in \textit{Discriminator}, we can obtain some output from intermediate layers, and the weighed sum of the delta L1 losses is defined as discrimination loss.}
\end{figure*}

In the goal above where cross-entropy loss function using $log$ criterion is a classic tactic. As Jianhua Lin explained in this paper\cite{kljs}, cross-entropy can be used to measure the Shannon Entropy needed to eliminate the uncertainty between two distributions, so it should naturally be the measurement of the disparity between two distributions P and Q:
\begin{align}
    H(P\|Q)=\mathbb{E}_{x\sim pX}[-\log Q(x)]=-\sum P(x)\log Q(x)
\end{align}
In machine learning field, cross-entropy is not the unique approach. Since during the training process of GAN, P can be seen as a constant variable which represents the distribution of normal sample. Hence Kullback-Leibler(KL) divergence also can be used to weigh the diversity of two distributions because:
\begin{align}
    KL(P\|Q)=H(P\|Q)-H(P)
\end{align}
Yet there is a deficiency in KL divergence: $KL(P\|Q)\ne KL(Q\|P)$, which means the KL divergence is asymmetrical so it can not be used to represent the distance between two distributions. This a fatal factor since inconsistent discrepancy brings no benefit to our training -- we do not know whether $KL(P\|Q)$ or $KL(Q\|P)$ should be taken to represent the gap between the two distributions. Thus Jensen-Shannon(JS) divergence is designed as follows to satisfy the symmetry required by distance:
\begin{align}
    JS(P\|Q)=\frac{1}{2}KL(P\|\frac{P+Q}{2})+\frac{1}{2}KL(Q\|\frac{P+Q}{2})
\end{align}
Profited from the duality in this principle, JS divergence can be seen as a sort of distance. Actually, JS divergence is the foundation most GANs work on and this divergence indeed helps a lot in image generation, in which field the dimension expansion is not so severe. Whereas for data with discrete features in cyber-intrusion detection such as the 0-1 representation of logic gate and non-numeric value that relies on \textit{One-Hot Encoding} or \textit{Dummy Encoding}, its dimension expansion can be very severe so that there will be few overlapping between real samples and generated samples. For example, when a sample from latent space is mapped into real sample space with higher dimension through \textit{Generator}, all the varieties in higher dimension space are actually be constrained by the sample from lower dimension space. So the dimension of the support of the higher dimension space is the dimension of the latent space in fact. Under the sway of dimension expansion, two distributions will have few overlapping inevitably, but discrete feature will aggravate it because of \textit{One-Hot Encoding} or \textit{Dummy Encoding}. And when two distribution have few overlapping, JS divergence will unavoidably converge to a constant, then leads to the happening of \textit{vanishing gradient}. Fortunately, in previous work, Martin Arjovsky et al.\cite{wgan} supplanted the JS divergence with Wasserstein Distance, which performs well even on the discrete distribution. Inspired by Wasserstein distance, we modify the training goal of our model as follows:
\begin{align}
    \min\limits_{G,E}\max\limits_{D}V(D,E,G)
\end{align}
\begin{align}
    V(D,E,G)=& \;\mathbb{E}_{x\sim pX} [\mathbb{E}_{z\sim pE(\cdot|x)}\| D(x,z)\|_{w}]\nonumber\\ &+ \mathbb{E}_{z\sim pZ} [\mathbb{E}_{x\sim pG(\cdot|z)}[1-\| D(x,z)\|_{w}]]
\end{align}
Where D, E, G separately represents \textit{discriminator}, \textit{encoder} and \textit{generator}, $pX$ is the distribution of normal samples $x$, $pZ$ is the distribution over the latent space, and $pE(z|x)$, $pG(x|z)$ is the distribution learned by \textit{encoder} and \textit{generator} respectively. $w$ represents the Wasserstein distance, it also facilitates the discrimination process compared with cross-entropy measurement and helps to ameliorate the generation process to generate more stable and more premium results.

In addition, the network should also be optimized. As is known to all, the data used to identify an intrusion is not like an image. In an image, where features around a specific feature usually have relevance with each other. For example, a pixel can be seen as a feature in image vector(image will be unrolled as a vector during training process), suppose there is a pixel in the canopy of a tree image, then the pixels surrounding it should intuitively outline the silhouette of the canopy. However, the features of the working status of a machine are independent features, there rarely exists relevance between those features. In this scenario, classical convolution kernel exerts poor influence on extracting features, so we adopt the FC(Full Connection) layer with dropout operation to construct our networks.

\subsection{Anomaly Assessment}\label{sect:anoass}
As for anomalous intrusion detection task, an evaluation standard is of necessity. Given there does not exist a unified criterion to assess the quality of generated result -- most applications of GAN were aimed to image generation so we can distinguish the superior or inferior through our naked eye, a principle is needed to guide us how to judge the quality of generated samples not similar with images. Since the greater the difference between the testing sample and the learned distribution of normal status, the more likely the testing sample is anomalous. Thus the discrepancy can be taken into account to evaluate the samples and there is no need to evaluate the result generated through \textit{generator} directly. We have tried several definitions of \textit{anomaly score} but they are essentially similar, inspired by definition proposed by Thomas Schlegl et al.\cite{anogan}, the \textit{anomaly score} we designed is as below:
\begin{align}
    S=(1-\sum_{i=1}^{n}\lambda_i)L_{R}+\sum_{i=1}^{n}\lambda_iL_{Di}
\end{align}
Where $\lambda_i$ is a constant and
\begin{align}
    L_{R}=|x-G(z)|
\end{align}
\begin{align}
    L_{Di}=\sum |f_{i}(x,z)-f_{i}(G(z),z)|
\end{align}
$S$ is the \textit{anomaly score}, $L_{R}$ is called \textit{Residual Loss}, used to measure the dissimilarity between testing sample and the regenerated sample, in this formula, $z$ and $x$ are corresponding point in latent space and real space respectively, under the postulation that a perfect \textit{generator} and a perfect mapping from latent space to real space, we have $L_{R}=0$ because $x$ and $G(z)$ are identical.
The second loss $L_{D}$ is defined as \textit{Discrimination Loss}, whose function is learning the feature representing. As emphasized by Goodfellow et al.\cite{itfg}, feature matching addresses the instability of GANs due to over-training on the \textit{discriminator} response, so in the feature matching technique, the \textit{generator} is mandated to generate data that has similar statistics as the training data instead of optimizing the parameters of the \textit{generator} by maximizing the output of \textit{discriminator} on generated examples(Eq.7). 

In \textit{Discrimination Loss}, $f$ is an intermediate layer embedded in \textit{discriminator}, $f(\cdot)$ is the output of this layer, the $\Sigma$ reveals there can exist several intermediate layers up to actual situation, and the closer the intermediate layer to the final logits produced by \textit{discriminator}, the coefficient should be greater. The multiple intermediate layers(Figure 4) help to better evaluate the difference between the pair of \textit{discriminator}'s input, with the introduce of intermediate layer, the adaptation of the coordinates of $z$ does not only rely on a hard decision from the trained \textit{discriminator}, about whether or not a generated image $G(z)$ fits the learned distribution of normal data, but also takes the rich information about the feature representation into account during the learning process of \textit{discriminator}. The $L1$ loss criterion of intermediate layers is also known as Feature Matching.\footnote{compared with the cross-entropy method in this paper\cite{egan}}

The last part is the investigation about how to select the anomalous sample according to their \textit{anomaly score}. We proposed two criterion
\begin{itemize}
    \item[$\bullet$] The first one is more practical in real life. In simple words, we need to add abundant already-known intrusion samples into a well-pretrained model to procure their \textit{anomaly score}. Empirically, we can find a threshold to determine the intrusion, in later detection we can judge a sample whether anomalous mainly from their \textit{anomaly score} -- less than the threshold represents normal, and vice versa. This method is proper to online detection, for there is no need to make sense of the proportion of normal samples and abnormal samples, all we need is a threshold obtained from experience.

    \item[$\bullet$] The second method is based on the proportion of normal samples and abnormal samples, this method is usually applied to the test on dataset and thus to evaluate the performance of the model. In practice, we need the contaminate rate\footnote{the empirical ratio of anomalous samples $:$ anomalous $/$ (normal $+$ anomalous)} $c\%$ before testing, after the \textit{anomaly score} of all the samples be computed, we take the top $c\%$ score and label their relative samples as anomalous intrusions.
\end{itemize}
\begin{table}
  \caption{Performance on KDD-99 dataset}
  \label{tab:freq}
  \begin{tabular}{cccc}
    \toprule
    Model&Precision&Recall&F1\\
    \midrule
    Isolation Forest &0.4415&0.3260&0.3750\\
    OC-SVM &0.7457&0.8523&0.7954\\
    DSEBM-r &0.8521&0.6472&0.7328\\
    DSEBM-e &0.8619&0.6446&0.7399\\
    $AnoGAN_{FM}$ &0.8786&0.8297&0.8865\\
    $BiGAN_{FM}$ &0.6578&0.7253&0.6899\\
    Our Model&\textbf{0.9324}&\textbf{0.9473}&\textbf{0.9398}\\
  \bottomrule
\end{tabular}
\end{table}
\section{Experiment}\label{sect:sample-experiment}
\subsection{Dataset}
Our experiment was based on KDD-99~(10 percent), a dataset widely used for the testing of cyber-intrusion detector. This database contains a standard set of data to be audited, which includes a wide variety of intrusions simulated in a military network environment. Each sample in this dataset is a network connection recording and has 41  features such as connection time, protocol type and a label noted as 'normal' or a certain attack name which represents 'abnormal'.

\subsection{Data Preprocessing}
Note that, in this dataset, the quantity of 'abnormal' samples far outnumber the 'normal' samples, which is incompatible with the actual situation where the 'normal' samples usually have the dominant quantity. Thus we follow the setup in this paper\cite{egan}, label the 'abnormal' samples as the 'normal' and the 'normal' samples as the 'abnormal'. This trick will not affect the identification ability of model because pure intrusion detection is a binary classification problem(anomalous intrusion or not). Accurately discern the normal status also means accurately discern the anomalous intrusion. In addition, for those discrete features whose value is not numeric, we recommend \textit{Dummy Encoding} or \textit{One-Hot Encoding}.

Before the training commences, we randomly dichotomize the initial dataset(around 500,000 samples) as two sets, then choose the normal-label samples from one set as the training set to train our model, subsequently pick the normal-label and abnormal-label samples from the other set in proportion to contaminate rate as the testing set.

\subsection{Results}
We reappeared several models including traditional anomaly detection methods such as Isolation Forest, Robust Covariance and previous GAN-based models for anomaly detection. The result of the comparison between them and our model is demonstrated in Table 1.\footnote{Values of $OC-SVM$, $DSEBM$ and $AnoGAN_{FM}$ are from paper\cite{egm}\cite{anogan}}\footnote{The result of $BiGAN_{FM}$ is obtained through the source code provided by the author\cite{egan}, all experiments ran on the same conditions, but the precision, recall and F1 pronounced in their paper was 0.8698, 0.9523 and 0.9058.}

Meanwhile, considering different occurrence frequency of cyber-intrusion, we also have explored the effect that different contaminate rate will exert on our model, we decreasingly choose $20\%$, $10\%$, $5\%$ and $1\%$ as the contaminate rate, the changes of precision, recall and F1 score are showed in Figure 5.
\begin{figure}
\includegraphics[width=8cm, height=4.6cm]{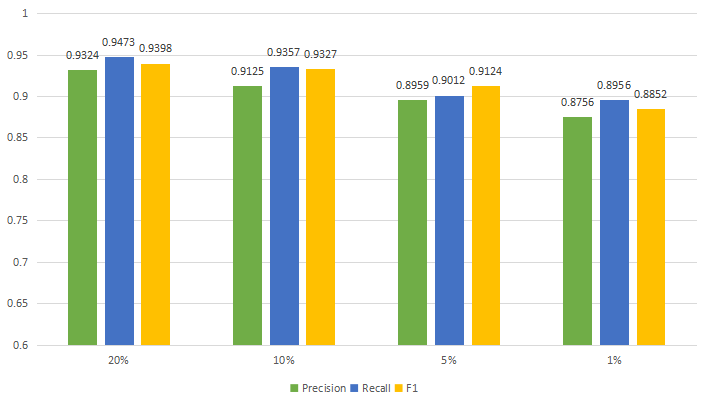}
\caption{Performance of our model on KDD-99 with 20\%, 10\%, 5\%, 1\% contaminate rate respectively.}
\vspace{-8pt}
\end{figure}
\subsection{Overhead}
As mentioned by Tim Bass\cite{h}, even a model can reach 100\% accuracy in detection, it should consider the detection latency, because the adversary still have enough time to damage the system if detection process cost excessive time.

In theory, during the training process, Wasserstein distance, which is demonstrated as the form of $L1$ loss, will cost less computation source than cross-entropy loss function. Meanwhile, during the testing time, when computing the \textit{anomaly score} of samples, our feature matching method in \textit{Discrimination loss} still performs better than the cross-entropy method. Given this GAN-based model\cite{egan} has the best performance on anomaly detection presently, we select it as the benchmark. The comparison of training time is showed in Figure 6. Besides, Figure 7 demonstrates the overhead of testing process. Since different configuration of a machine will lead to a heterogeneous testing result, we separately show the comparison in Figure 7: (a) is the comparison between AnoGAN and BiGAN, based on NIVIDA Tesla K40 GPUs, pronounced by Houssam Zenati et al.\cite{egan}. (b) is the comparison between BiGAN\cite{egan} and our model based on Intel(R) Core(TM) i5-5200U CPUs. From Figure 6 and Figure 7, it can be clearly seen that our model has better performance on training or testing process compared with previous GAN-based model.

\begin{figure}
\includegraphics[width=7cm, height=3.6cm]{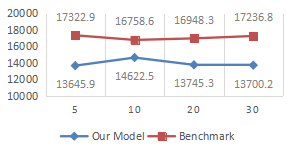}
\caption{Average Training Time(s) of several experiments ran on Intel(R) Core(TM) i5-5200U CPUs. Benchmark is the state-of-the-art GAN-based model on anomaly detection\cite{egan}.}
\end{figure}

\begin{figure}[htp]
\centering
\begin{minipage}[b]{0.3\textwidth}
\centering
\subcaption{NIVIDA Tesla K40 GPUs, Tensorflow 1.1.0 and Python 3.5.3}
\begin{tabular}{ccc}
\toprule
Model&Time(ms)\\
\midrule
$AnoGAN_{FM}$&3527\\
$BiGAN_{FM}$&5.3\\
\bottomrule
Speed Up&$\sim$660\\
\bottomrule
\end{tabular}
\end{minipage}
\begin{minipage}{0.3\textwidth}  
\centering
\subcaption{Intel(R) Core(TM) i5-5200U CPUs, Tensorflow 1.1.0 and Python 3.5.3}
\begin{tabular}{ccc}
\toprule
Model&Time(ms)\\
\midrule
$BiGAN_{FM}$&1.9\\
Our Model&1.4\\
\bottomrule
Speed Up&$\sim$1.357\\
\bottomrule
\end{tabular}
\end{minipage}
\centering
\caption{Average testing time over 100 batches}
\end{figure}
\section{Conclusion}\label{sect:sample-conclusion}
We demonstrated our GAN-based model can be used for cyber-intrusion detection task which enable the effective recognition of anomalies on unknown data based on unsupervised training. In general, the bilateral transformation structure is conducive to constructing the mapping between latent space and real space in a more accurate way, the Wasserstein distance we adopted performs well on weighing the disparity between two rarely overlapping distributions and the \textit{multiple intermediate layers} are advantageous to appraise the \textit{anomaly score} of a targeted sample. The model we designed outperforms previous GAN-based models on cyber-intrusion detection task. Meanwhile, it remarkably curtails the time cost on training and testing process.

In future work, we plan to investigate the temporal influence on cyber-intrusion detection. The occurrence of some intrusions may be owing to a chronic process, in which situation we can not discern a delitescent anomaly instantly but must wait for a period of time. So if the ability to dispose of temporal feature can be inset in GAN, the model is feasible to solve intrusion detection problems in a wider range.

\bibliographystyle{ACM-Reference-Format}
\bibliography{sample-bibliography}

\end{document}